\documentclass[runningheads]{llncs}
\usepackage[T1]{fontenc}
\usepackage{graphicx}

\usepackage{algorithm}
\usepackage{algpseudocode}
\usepackage{dsfont}
\usepackage{amsmath, amssymb}
\usepackage{bbm}
\usepackage{subcaption}
\usepackage{float}
\usepackage{bm}
\begin{document}
\title{K-Survival Means}
\author{Abdallah Alabdallah\orcidID{0000-0001-9416-5647} 
% \and
% Third Author\inst{3}\orcidID{2222--3333-4444-5555}
}
\authorrunning{A. Alabdallah}
\institute{Center for Applied Intelligent Systems Research (CAISR), Halmstad, Sweden \\
\email{abdallah.alabdallah@hh.se}\\
}
% \author{Abdallah Alabdallah\orcidID{0000-0001-9416-5647} 
% % \and
% % Second Author\inst{2,3}\orcidID{1111-2222-3333-4444} \and
% % Third Author\inst{3}\orcidID{2222--3333-4444-5555}
% }
% %
% \authorrunning{A. Alabdallah}
% \institute{Center for Applied Intelligent Systems Research (CAISR), Halmstad, Sweden
% \email{abdallah.alabdallah@hh.se}\\
% }
%==================================================
%
\maketitle              % typeset the header of the contribution
\begin{abstract}
In this work, we propose K-SurvMeans, a novel extension of K-Means for clustering survival data. The method explicitly uses the survival outcome in the clustering process to optimize cluster centers, thereby maximizing pairwise survival differences between clusters. The objective function encourages the clusters to be well-separated from the survival perspective. Since the resulting optimization problem is non-differentiable, we employ the Particle Swarm algorithm for the Optimization process.

To further improve flexibility and mitigate the curse of dimensionality, we extend the framework to operate in a learned low-dimensional latent space obtained via a dimensionality reduction. This allows the method to capture better-separated clusters and enhance optimization efficiency by reducing the search space.

Experiments on multiple publicly available benchmark survival datasets demonstrate that K-SurvMeans consistently yields clusters with improved separation in survival distributions compared to existing deep learning-based survival clustering methods. Full implementation is available on GitHub~\footnote{https://github.com/abdoush/KSurvivalMeans}

\keywords{K-Means  \and Survival Analysis \and Clustering.}
\end{abstract}
\section{Introduction}

Survival analysis is a branch of statistics concerned with modeling and analyzing the time until the occurrence of an event of interest. It originated in health care and medical studies to estimate the time to critical events in patients, such as mortality, disease progression, and treatment response. However, survival analysis applications extend to various fields, including predictive maintenance, reliability engineering, customer churn analysis, and financial risk assessment.

In healthcare, patient stratification is a fundamental task that enables clinicians to identify patient groups with distinct risk profiles, support personalized intervention and treatment planning, and more efficiently allocate healthcare resources~\cite{Aljohani2024}. Similarly, in clinical studies, population heterogeneity often poses a significant challenge, as the presence of subgroups with different prognoses may compromise the performance and interpretability of predictive models~\cite{Alabdallah2022survshap}.

Clustering techniques offer a natural approach for discovering patient subgroups and identifying populations with different risk characteristics. However, conventional clustering algorithms, such as K-Means, operate mainly on the input features and seek groups of observations that are similar in the feature space. While such clustering is useful for uncovering patterns in the data, these methods do not explicitly consider survival outcomes during the clustering process. Consequently, the resulting clusters may exhibit similar survival distributions, limiting their usefulness for risk stratification and survival-related decision-making.

Due to the importance of survival analysis in health care applications and the increasing amount of survival data collected over the years, a wide range of machine learning models has been developed and adapted to survival outcome estimation~\cite{Ishwaran_2008,Katzman:2018,Lee_2018,Kvamme:2019,ALABDALLAH2024surved,ALABDALLAH2026coxse}. In contrast, clustering methods specifically designed for survival analysis remain relatively underexplored. Few existing studies have primarily focused on deep learning approaches that learn latent representations jointly optimized for clustering and survival prediction. By incorporating survival information into the representation learning process, these methods aim to produce clustered latent representations with distinct survival characteristics. However, little attention has been devoted to extending traditional clustering algorithms to survival analysis, despite their simplicity, interpretability, and widespread adoption. Notably, algorithms such as K-Means are frequently employed as baseline methods in survival clustering studies, yet they do not leverage survival information during optimization.

In this work, we introduce K-SurvMeans, a novel extension of K-Means for survival data clustering. Unlike conventional K-Means, the proposed method explicitly incorporates survival outcomes into the clustering objective, guiding the optimization process to learn clusters that exhibit statistically significant differences in survival behavior. The resulting clusters provide a meaningful stratification of individuals according to their survival profiles, thereby enhancing the utility of such clustering for survival analysis and risk-based decision-making.

% K-Means Survey:
% \cite{IKOTUN2023178}

% k-Means: \cite{steinhaus1956division,jancey1966multi,macqueen1967multivariate,lloyd1982least}

\section{Survival Analysis Background}
Survival analysis is concerned with modeling the time until an event of interest, such as patient death or system failure. A major challenge motivating its development is the presence of incomplete outcome information, where, at the end of a study, some subjects might not have yet experienced the event. Such observations are referred to as censored data, since their exact event times are unobserved. Survival data are typically represented as triplets $(\mathbf{x}_i, t_i, e_i)$, where $\mathbf{x}_i$ denotes the feature vector of subject $i$, $t_i$ denotes the observed follow-up time, and $e_i \in \{0,1\}$ is a binary event indicator specifying whether $t_i$ corresponds to the event time ($e_i = 1$) or a censoring time ($e_i = 0$). 

Unlike other machine learning methods that predict a point estimate, like regression or classification, most survival models estimate functions. The main function that survival models aim to estimate is the survival function, where for a random variable $T$, the survival function is the probability of surviving beyond time $t$ and is defined as:
\begin{equation}
    S(t) = P(T > t)
    \label{eq:survival_finction}
\end{equation}
The Kaplan–Meier estimator~\cite{Kaplan:1958} is the earliest and most widely used non-parametric method for estimating the survival function $S(t)$, and is defined as:
\begin{equation}
\hat{S}(t)=\prod_{i:t_i \leq t} \left( 1 - \frac{d_i}{n_i} \right),
\label{eq:KMS}
\end{equation}
where $d_i$ is the number of events that happened at time $t_i$, and $n_i$ is the number of the individuals at risk at time $t_i$. The Kaplan–Meier estimator is a population-level estimator that doesn't rely on the subjects' features. However, in later years, many models were proposed to condition predictions on explanatory features to provide personalized survival function predictions, most notably the Cox Proportional Hazards model ~\cite{Cox1972}, which assumes that the explanatory features have an exponential multiplicative effect on a baseline hazard function; a closely related function to the survival function. Cox formulates the hazard function as:
\begin{equation}
    h(t, \bm{x}) = h_0(t) e^{\bm{w}^{\intercal} \bm{x}}
    \label{eq:cph_ht}
\end{equation}

More recently, with the advancement of machine learning, more advanced models were introduced relying on machine learning~\cite{Ishwaran_2008,VanBelle_2009}, and deep learning~\cite{Katzman:2018,Lee_2018,Kvamme:2019,ALABDALLAH2024surved,ALABDALLAH2026coxse} models. 

\section{Related Work}

Clustering of survival data aims to identify subpopulations exhibiting distinct survival patterns within a heterogeneous population. Compared to traditional survival prediction, which focuses on estimating individual risk scores or time-to-event distributions, survival clustering seeks to uncover hidden groups that are characterized by different time-to-event outcomes. 

Earlier attempts to cluster survival data either relied on the feature space only~\cite{ahlqvist2018novel} or indirectly incorporated survival outcomes in clustering through feature-selection techniques to select features with high correlation with the clinical variable of interest~\cite{Bair2004}. Recently, deep learning techniques started to be used for clustering survival data. Most notably, among the earliest deep learning approaches to this problem is Survival Cluster Analysis (SCA) proposed by ~\cite{chapfuwa2020survival}. SCA models population heterogeneity through a Bayesian nonparametric framework that represents individuals in a clustered latent space while encouraging both accurate time-to-event prediction and the discovery of subpopulations with distinct risk profiles. More recently, Deep Variational Approach to Clustering Survival Data (VaDeSC) \cite{ManduchiMarcinkevics2022} has been proposed. VaDeSC is a semi-supervised probabilistic model based on variational inference that extends deep generative clustering~\cite{Jiang2017p273,dilokthanakul2017} to survival analysis by jointly modeling explanatory variables and censored survival outcomes through a variational autoencoder, enabling the discovery of clusters associated with distinct survival distributions. These methods share a common characteristic: they rely on deep latent-variable models that jointly learn feature representations, cluster assignments, and survival-related objectives. While highly expressive, these methods focus on deep learning-based models, which often introduce substantial model complexity, require larger datasets, and require extensive hyperparameter tuning. Furthermore, little attention has been paid to traditional clustering algorithms, which are relatively unexplored in the context of survival data. 

While classical clustering methods such as K-Means remain widely used as baseline approaches in survival clustering studies due to their simplicity, conventional K-Means optimizes cluster compactness in the feature space without considering survival outcomes, which may result in clusters exhibiting similar survival behavior. The proposed K-SurvMeans addresses this gap by extending K-Means with a survival-driven optimization objective based on pairwise log-rank statistic~\cite{Peto_1972}. K-SurvMeans directly searches for cluster centers that maximize survival separation between clusters, thereby retaining the simplicity and interpretability of centroid-based clustering while incorporating survival information into the optimization process.

\section{Method}
% K-SurvMeans extends the K-Means algorithm to the realm of survival analysis. The objective is to find $K$ clusters with significantly different survival behaviors. 

% Given a dataset $X \in \textsc{R}^{n \times p}$ consisting of $n$ observations $\bm{x} \in \textsc{R}^{p}$ and a terget variable $y$ consisting of a tuple $(t, e)$ where $t \in \textsc{R}$ is the censored or uncensored event time and $e \in {0, 1}$ is the event indicator that takes the value $1$ if the event happend at time $t$ and takes the value $0$ otherwise. 

% The objective is to find $k$ centers $c_i \in \textsc{R}^p: i \in \{0, 1, .., k-1\}$ that maximize the pairwise differences based on the log-rank statistical test. 

K-SurvMeans extends the K-Means clustering algorithm to survival analysis. Its objective is to partition the data into \(k\) clusters exhibiting maximally distinct survival patterns.

Let \(X \in \mathbb{R}^{n \times p}\) denote a dataset comprising \(n\) observations \(\mathbf{x} \in \mathbb{R}^{p}\). Associated with each observation is a survival outcome \(y=(t,e)\), where \(t \in \mathbb{R}^{+}\) represents the observed event or censoring time, and \(e \in \{0,1\}\) is the event indicator, taking the value \(1\) if the event occurred at time \(t\) and \(0\) if the event did not happen until after time \(t\), in which case the observation is called right-censored.

The goal is to determine a set of \(k\) cluster centers,
\[
C=\{c_0,c_1,\ldots,c_{k-1}\}, \qquad c_i \in \mathbb{R}^{p},
\]
such that the resulting partition maximizes the separation between clusters with respect to their survival distributions. To quantify this separation, pairwise comparisons between clusters are performed using the log-rank test, and the cluster centers are optimized to maximize the overall pairwise survival differences.

% Given a set of $k$ clusters $C=\{c_0, c_1, .., c_{k-1}\}$, a mapping function that maps each observation in the data to its cluster:

% \begin{math}
%     \textsc{g}(c_i)=\{x \in X}
% \end{math}

Given a set of $k$ cluster centers $C$, let
\[
g:X \rightarrow C
\]
denote the cluster assignment function, where $g(x)=c_i$ indicates that observation $x\in X$ is assigned to cluster center $c_i$. The set of observations assigned to $c_i$ is given by
\[
g^{-1}(c_i)=\{x\in X \mid g(x)=c_i\}.
\]

For each pair of clusters $(c_i,c_j)$, let
\[
(p_{ij},\, \chi^2_{ij})
=
\operatorname{log-rank}\!\left(g^{-1}(c_i), g^{-1}(c_j)\right),
\]
where $p_{ij}$ is the p-value and $\chi^2_{ij}$ is the corresponding log-rank test statistic.

Let $\alpha$ denote a predefined confidence threshold, and $\mathbbm{1}(.)$ the indicator function. We define the loss function as
\begin{equation}
\mathcal{L}(C; X, k)
= -
\left(
\sum_{0 \le i < j \le k-1}
\mathbbm{1}\!\left(
p_{ij} < \alpha
\right)
\right)
%\cdot
\left(
\sum_{0 \le i < j \le k-1}
\chi^2_{ij}
\right)
\label{eq:loss}
\end{equation}
We seek the optimal cluster centers given by
\[
C^{*} = \arg\min_{C} \mathcal{L}(C; X, k).
\]
To optimize the objective function, we employ Particle Swarm Optimization (PSO). In the proposed formulation, each particle encodes a candidate clustering solution as the concatenation of the coordinates of all cluster centers. Specifically, for a clustering configuration comprising k cluster centers in a p-dimensional feature space, a particle is represented by a vector of length $k \times d$. During the optimization process, particles iteratively update their positions and velocities according to the standard PSO update rules, balancing exploration and exploitation through the influence of their personal best positions and the global best position found by the swarm. For each particle, observations are assigned to their nearest cluster center. The resulting clusters are evaluated using the objective function defined in Equation~\ref{eq:loss}, and the resulting objective value is used as the particle's fitness score. The optimization terminates when a predefined stopping criterion is met, such as reaching a maximum number of iterations or achieving swarm convergence. 

%TODO: introduce latent representation
In this formulation, the dimensionality of each particle increases linearly with the number of clusters, thereby increasing the complexity of the optimization problem. Moreover, the dimensionality of the input data plays a critical role in this scaling behaviour. To address this issue, we introduce a dimensionality reduction function $f_\theta: \mathbb{R}^p \rightarrow \mathbb{R}^q$, which maps the original input space to a lower-dimensional latent representation $z \in \mathbb{R}^q$, where $q \ll p$.

In the current formulation, the mapping $f_\theta$ is learned independently of the optimization loop and is therefore fixed during clustering. In the simplest case, $f_\theta(x)=x$, which reduces the method to the original formulation in the input space. The overall procedure is summarized in Algorithm~\ref{alg:pso-ksurvmeans-latent}.

% In this formulation, the particle length increases with the number of clusters, thereby increasing the complexity of the optimization task. Moreover, the data's dimensionality plays a crucial role in the rate of such an increase. We introduce a dimensionality reduction function  $f_\theta: \mathbb{R}^p \rightarrow \mathbb{R}^q$ that learns a lower-dimensional latent representation $z \in \mathbb{R}^q$, where $q \ll p$. Currently, $f_{\theta}$ is not learned during the optimization loop; it is pretrained prior to the loop. In the simplest case, $f_{\theta}(x)=x$ brings back the original formulation. The algorithm is detailed in Algorithm~\ref{alg:pso-ksurvmeans-latent}

\begin{algorithm}[t]
\caption{K-SurvMeans}
\label{alg:pso-ksurvmeans-latent}
\begin{algorithmic}[1]

\State \textbf{Input:} Dataset $X$, survival outcomes $y$, number of clusters $k$, encoder $f_\phi$, swarm size $M$
\State \textbf{Output:} Cluster centers $C^*$ in latent space

\State Learn or initialize encoder $f_\theta: \mathbb{R}^p \rightarrow \mathbb{R}^q$, where $q \ll p$
\State Compute latent representations $Z = f_\theta(X)$

\State Initialize swarm of particles $\{\mathbf{z}_m\}_{m=1}^{M}$
\Comment{Each particle encodes $k$ cluster centers in $\mathbb{R}^{k \cdot q}$}

\For{each particle $m$}
    \State Randomly initialize $\mathbf{z}_m$
    \State Decode $\mathbf{z}_m \rightarrow C_m$
    \State Assign each $z \in Z$ to nearest center in $C_m$
    \State Evaluate fitness $\mathcal{L}(C_m; Z, y)$
    \State Set personal best $pbest_m \gets \mathbf{z}_m$
\EndFor

\State Set global best $gbest$ among all particles

\While{stopping criterion not met}
    \For{each particle $m$}
        \State Update particle position using PSO update rule
        \State Decode $\mathbf{z}_m \rightarrow C_m$
        \State Assign each $z \in Z$ to nearest center in $C_m$
        \State Compute fitness $\mathcal{L}(C_m; Z, y)$
        
        \If{$\mathcal{L}(C_m)$ better than $pbest_m$}
            \State Update $pbest_m \gets \mathbf{z}_m$
        \EndIf
    \EndFor

    \State Update $gbest$ among all particles
\EndWhile

\State \Return Cluster centers corresponding to $gbest$

\end{algorithmic}
\end{algorithm}

\section{Results and Discussion}
In this work, we compare six models. Two variants of KSurvMeans, one without dimensionality reduction, which we call KSurvMeans, and the other one with dimensionality reduction, which we term KSurvMeans(Latent). Similarly, we compare with a regular K-Means algorithm and another variant with dimensionality reduction, which we term KMeans and KMeans(latent). In both cases, we use PCA as a dimensionality reduction method with a fixed number of principal components set to $5$. We also compared with SCA and VaDeSC, two deep-learning-based approaches. For the comparisons, we used the percentage of pairwise significant differences across all pairs of clusters as the evaluation metric.

\begin{equation}
\mathcal{P}(k,C,\alpha)
= \frac{\sum_{0 \le i < j \le k-1} \mathbbm{1}\!\left(p_{ij} < \alpha \right)}{\binom{k}{2}}
\nonumber
\end{equation}
where $p_{ij}$ is the p-value associated with log-rank test between the clusters $c_j$ and $c_j$.

\subsection{Performance Comparison Across Varying Numbers of Clusters}
\begin{figure}[ht]
\centering
\begin{subfigure}{0.50\linewidth}
  \centering
  \includegraphics[width=1\linewidth]{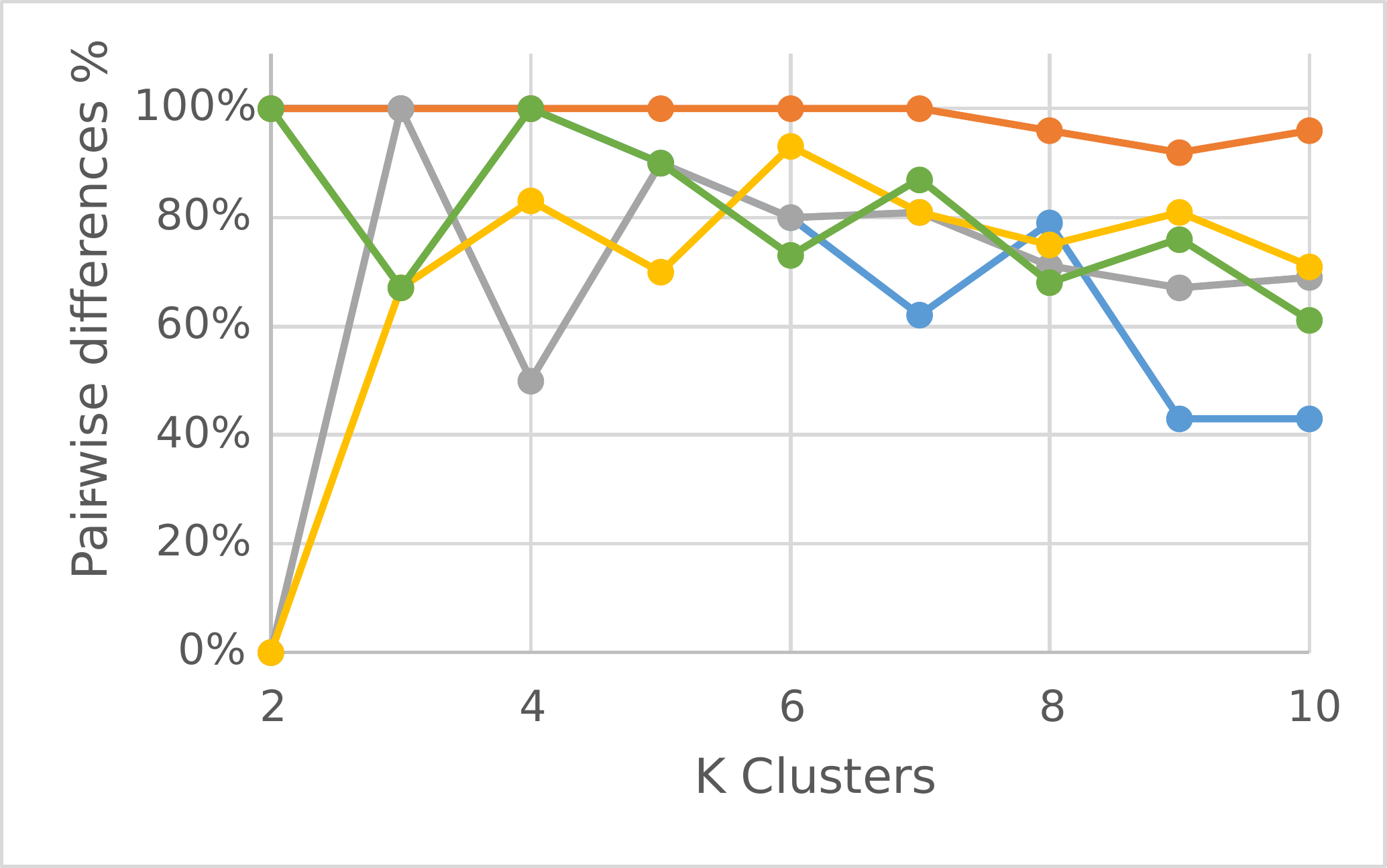}
  \caption{Flchain Dataset }
  \label{fig:stability_Lin}
\end{subfigure}%
~
\begin{subfigure}{0.50\linewidth}
  \centering
  \includegraphics[width=1\linewidth]{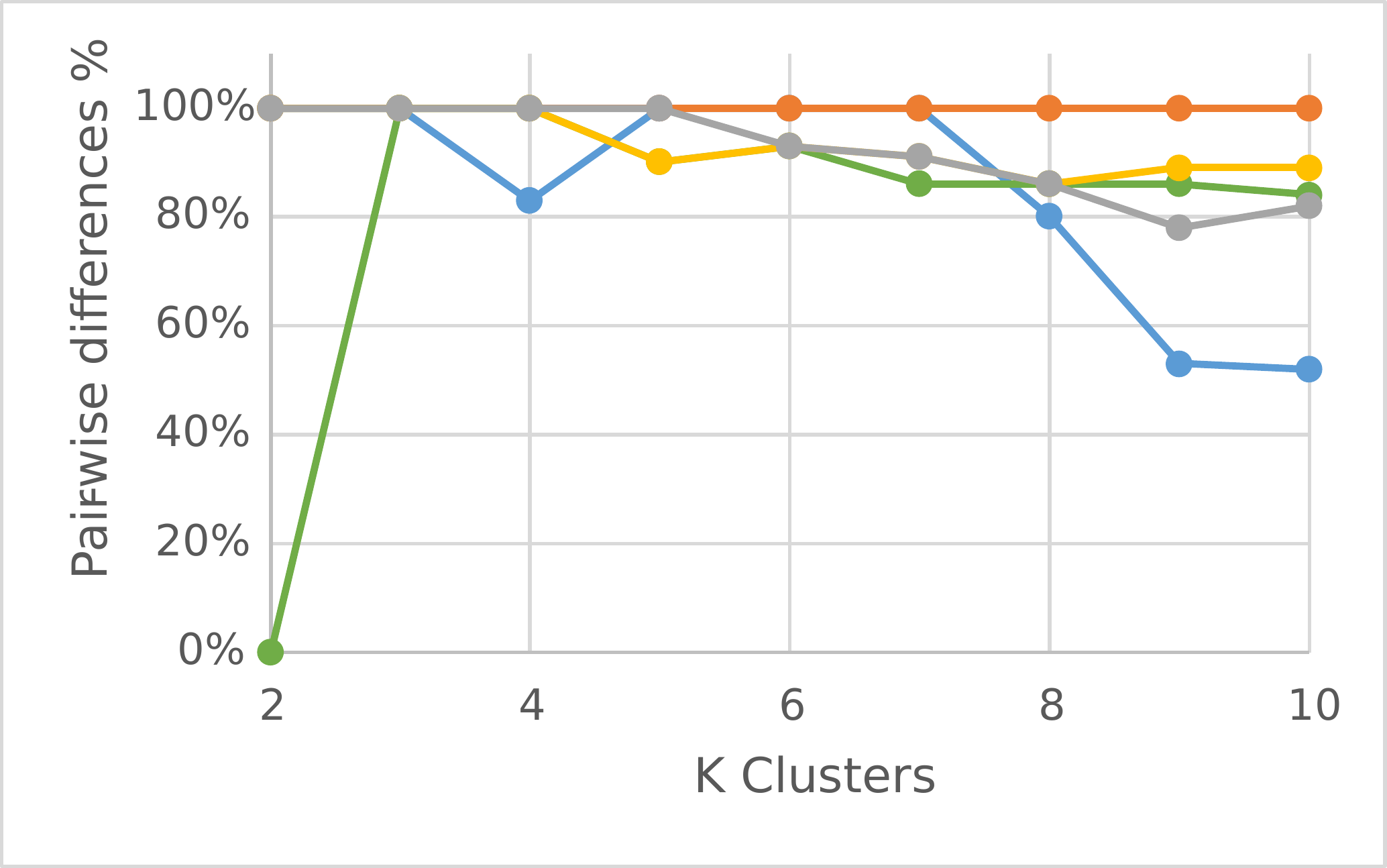}
  \caption{Support Dataset}
  \label{fig:stability_NonLin}
\end{subfigure}%

\begin{subfigure}{0.50\linewidth}
  \centering
  \includegraphics[width=1\linewidth]{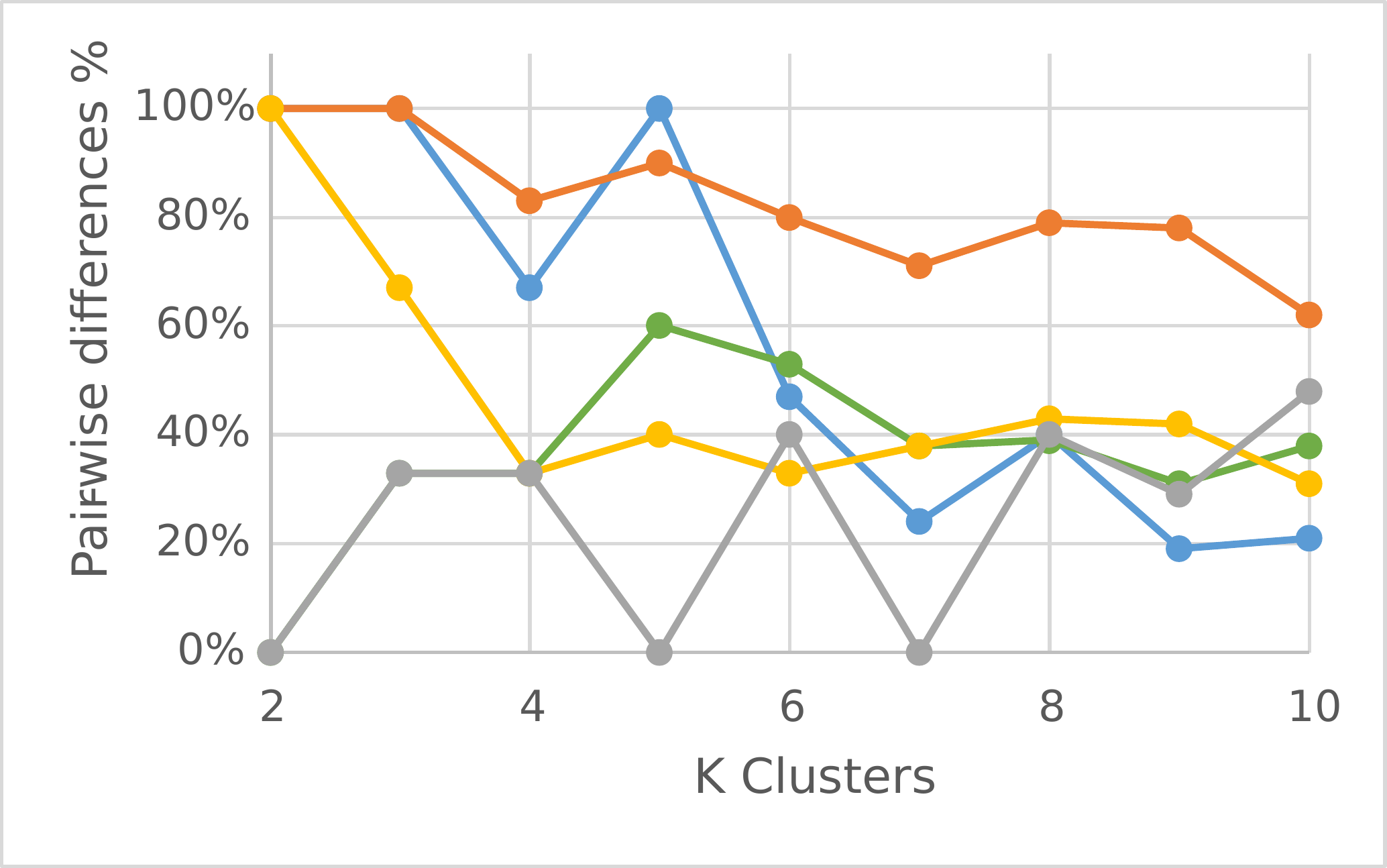}
  \caption{Metabric Dataset}
  \label{fig:stability_NonLinX}
\end{subfigure}%
~
\begin{subfigure}{0.50\linewidth}
  \centering
  \includegraphics[width=1\linewidth]{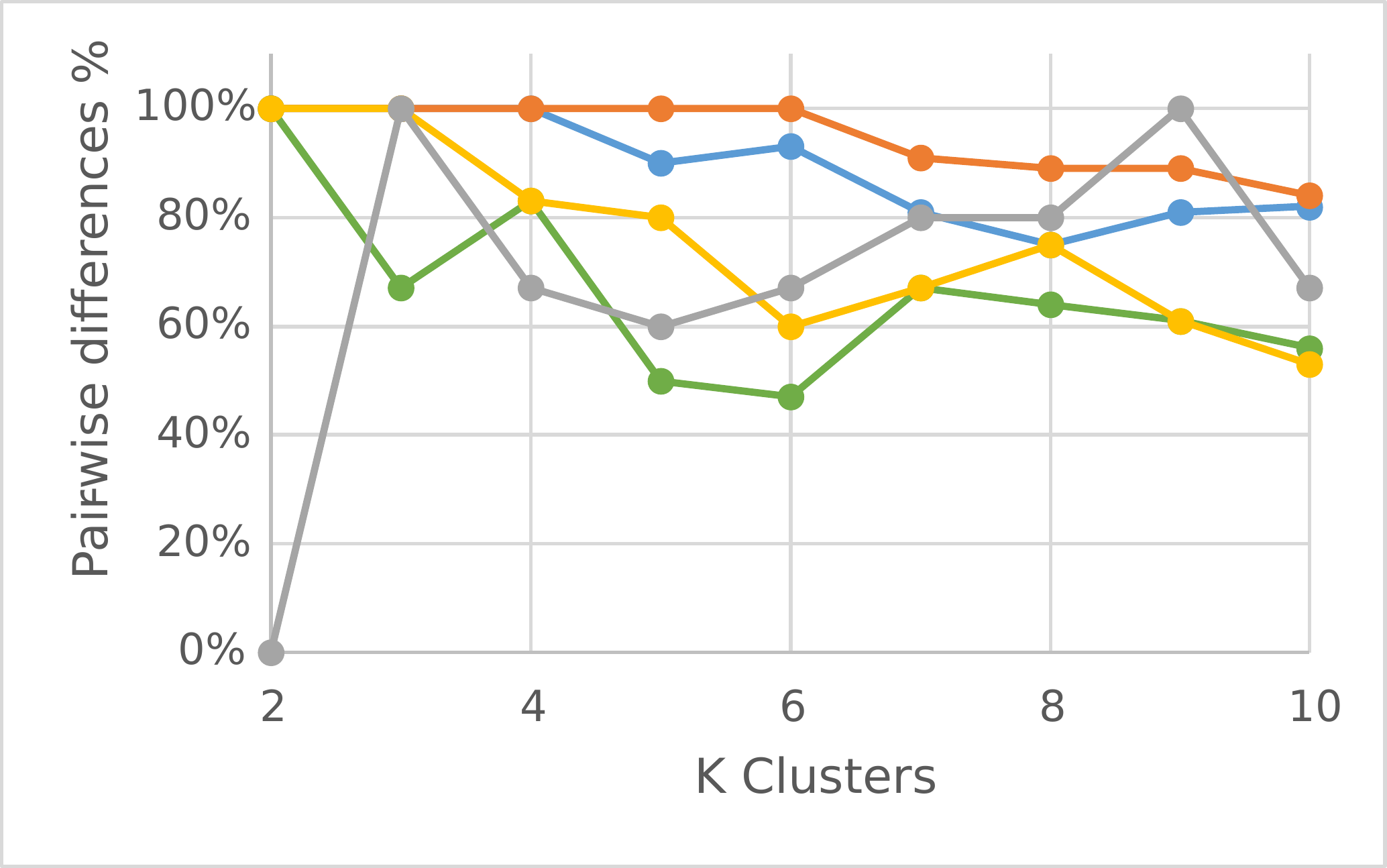}
  \caption{Nwtco Dataset}
  \label{fig:stability_Flchain}
\end{subfigure}%

\begin{subfigure}{1\linewidth}
  \centering
  \includegraphics[width=1\linewidth]{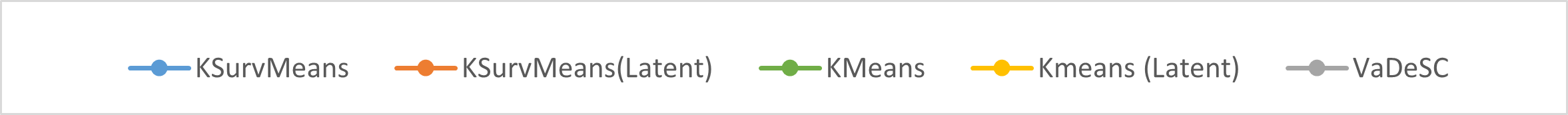}
  %\caption{SEER Dataset}
  \label{fig:stability_SEER}
\end{subfigure}
\caption{Percentage of significant pairwise cluster differences across varying numbers of clusters for different models and datasets.}
\label{fig:performance_vs_k}
%\vspace{-2em}
\end{figure}

In this section, we compare the performance of the models that depend on the choice of the number of clusters $k$. We excluded SCA as it doesn't require specifying the number of clusters. The results depicted in Figure~\ref{fig:performance_vs_k} show that KSurvMeans(Latent) showed the best performance across the four datasets, where it was able to identify clusters with the highest percentage of differences across various cluster numbers. Interestingly, KSurvMeans showed high performance with a lower number of clusters, which degrades with the increasing number of clusters. This can be attributed to the particle's increased dimensionality, which makes optimization harder.

%\newpage
\subsection{Generalization Performance}
In this experiment, we compare the predictive performance of the models for clustering new, unseen data. For all the models except the SCA, we use a validation set to select the $k$ value corresponding to the model with the largest number of clusters with the highest percentage of pairwise differences in terms of the log-rank test. The SCA doesn't require the $k$ parameter and results in one model with an effective number of clusters. We report the results of all the models on a hold-out test set. The results reported in Table~\ref{tab:gen_performance} list the number of clusters found by each model along with the percentage of significant pair-wise differences between the found clusters. For the SCA method, we use the term (effective x) to refer to the number of clusters found by the algorithm on the test set. In some cases, the VaDeSC algorithm converges to a lower number of clusters than the predefined k, in which case we also use the term (effective x).

\begin{table}[H]
\caption{Comparison of the models' performances. The cell values represent $n (p\%)$, where $n$ is the number of clusters, and $(p\%)$ is the percentage of significant pair-wise differences.}\label{tab:gen_performance}
\resizebox{\textwidth}{!}{%
\begin{tabular}{|l|c|c|c|c|}
\hline
Model &  FLCHAIN & SUPPORT & METABRIC & NWTCO\\
\hline
KSuvMeans          & 3 (100\%) & 2 (100\%) & 2 (100\%) & \textbf{2 (100\%)} \\
KSuvMeans (Latent) & \textbf{5 (100\%)} & \textbf{5 (100\%)} & \textbf{3 (100\%)} & \textbf{2 (100\%)} \\
KMeans             & 3 (100\%) & 3 (100\%) & 2 (100\%) &  \textbf{2 (100\%)}\\
KMeans (Latent)    & 10 (69\%) & 4 (100\%) & 2 (100\%) & \textbf{2 (100\%)} \\
SCA                & (effective 6) (33\%) & (effective 11) (53\%) & (effective 9) (19\%) & (effective 15) (42\%) \\
VaDeSC             & 2 (100\%) & 2 (100\%) & 5 (effective 3) (33\%) & 9 (effective 4) (83\%) \\
\hline
\end{tabular}
}
\end{table}

% The results show that K-SurvMeans and K-SurvMeans(Latent) have consistently achieved the highest percentage of differences between clusters; however, K-SurvMeans(Latent) found a larger number of clusters. Moreover, K-Means and K-Means(Latent) mostly achieved a high percentage of differences among clusters; however, with fewer clusters discovered. On the other hand, deep learning approaches tend to find a larger number of clusters with a smaller percentage of differences. Such well-cluster separation shown by K-SurvMeans(latent) is also evident in the Kaplan-Meier survival curves of the found clusters depicted in Figure~\ref{fig:survival_curves}.

The results show that K-SurvMeans and K-SurvMeans(Latent) have consistently achieved the highest percentage of differences between clusters; however, K-SurvMeans(Latent) found a larger number of clusters. This suggests that the proposed optimization objective is effective in identifying groups with distinct survival behaviors. Among the two variants, K-SurvMeans(Latent) generally identified a larger number of clusters while maintaining a high degree of survival separation, which demonstrates the benefit of performing the optimization in a lower-dimensional latent space. The latent representation appears to facilitate the discovery of finer-grained patient subgroups without compromising the distinction between their survival distributions.

Interestingly, the baseline K-Means and K-Means(Latent) approaches also achieved relatively high percentages of significant pairwise differences among the discovered clusters, despite not explicitly optimizing for survival separation. However, these methods typically produced fewer clusters. While the resulting clusters were often well-separated in terms of survival outcomes, they captured a lower level of population heterogeneity than K-SurvMeans(Latent).

In contrast, the deep learning-based methods tended to identify a larger number of clusters, suggesting a greater ability to capture complex structures within the data. Nevertheless, the survival differences between these clusters were generally weaker, as reflected by the lower percentage of significant pairwise differences. This observation indicates that increasing the number of clusters does not necessarily lead to more clinically meaningful stratifications if the resulting groups exhibit similar survival patterns.

\begin{figure}[H]
\centering

%==================== COLUMN HEADERS ====================

\begin{minipage}[c]{0.04\linewidth}
\end{minipage}
\hfill
\begin{minipage}[c]{0.23\linewidth}
    \centering \tiny\textbf{Flchain}
\end{minipage}
\hfill
\begin{minipage}[c]{0.23\linewidth}
    \centering \tiny\textbf{Support}
\end{minipage}
\hfill
\begin{minipage}[c]{0.23\linewidth}
    \centering \tiny\textbf{Metabric}
\end{minipage}
\hfill
\begin{minipage}[c]{0.23\linewidth}
    \centering \tiny\textbf{Nwtco}
\end{minipage}

\vspace{3mm}

%==================== ROW 1: K-SurvMeans ====================

\begin{minipage}[c]{0.04\linewidth}
    \centering
    \rotatebox{90}{\tiny\textbf{K-SurvMeans}}
\end{minipage}
\hfill
\begin{minipage}[c]{0.23\linewidth}
    \centering
    \includegraphics[width=\linewidth]{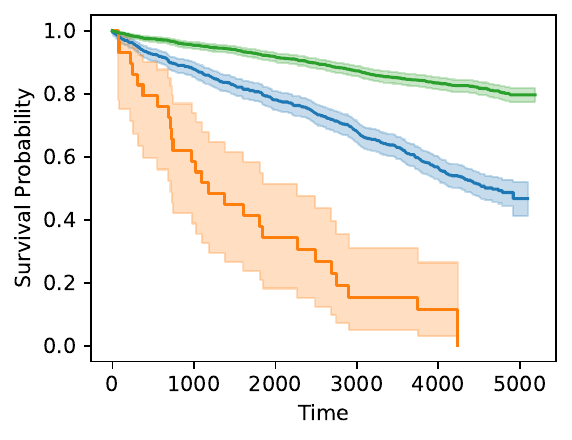}
    %\subcaption{K-SurvMeans}
\end{minipage}
\hfill
\begin{minipage}[c]{0.23\linewidth}
    \centering
    \includegraphics[width=\linewidth]{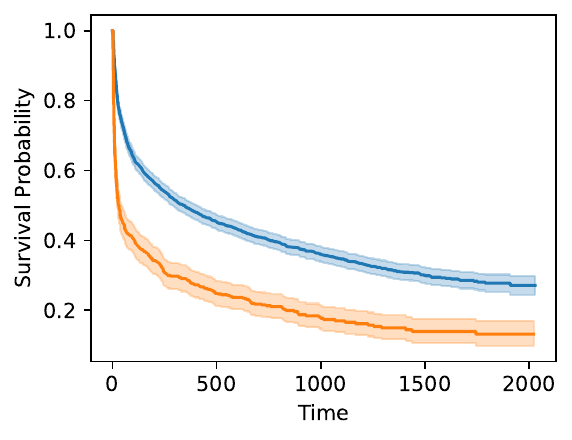}
\end{minipage}
\hfill
\begin{minipage}[c]{0.23\linewidth}
    \centering
    \includegraphics[width=\linewidth]{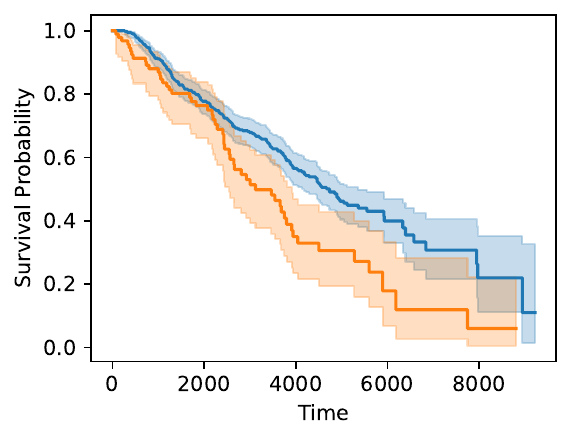}
\end{minipage}
\hfill
\begin{minipage}[c]{0.23\linewidth}
    \centering
    \includegraphics[width=\linewidth]{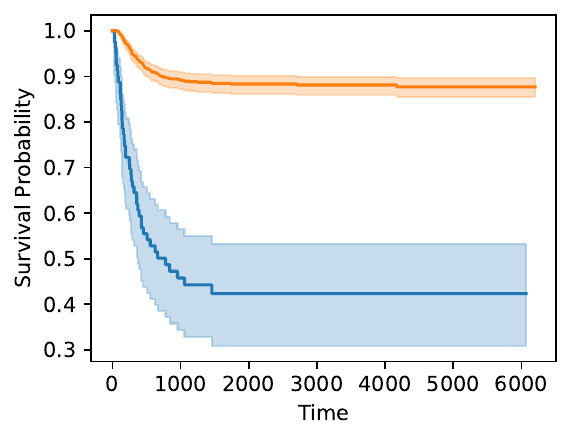}
\end{minipage}

\vspace{2mm}

%==================== ROW 2: Latent K-SurvMeans ====================

\begin{minipage}[c]{0.04\linewidth}
    \centering
    \rotatebox{90}{\tiny\textbf{K-SurvMeans(Latent)}}
\end{minipage}
\hfill
\begin{minipage}[c]{0.23\linewidth}
    \centering
    \includegraphics[width=\linewidth]{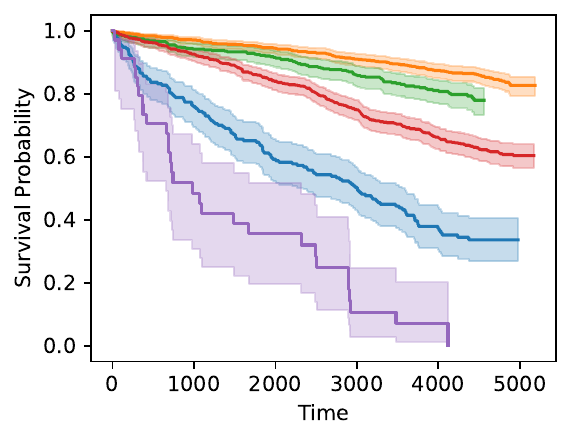}
\end{minipage}
\hfill
\begin{minipage}[c]{0.23\linewidth}
    \centering
    \includegraphics[width=\linewidth]{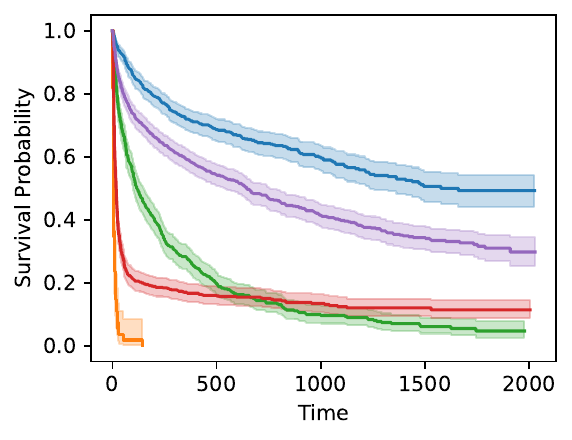}
\end{minipage}
\hfill
\begin{minipage}[c]{0.23\linewidth}
    \centering
    \includegraphics[width=\linewidth]{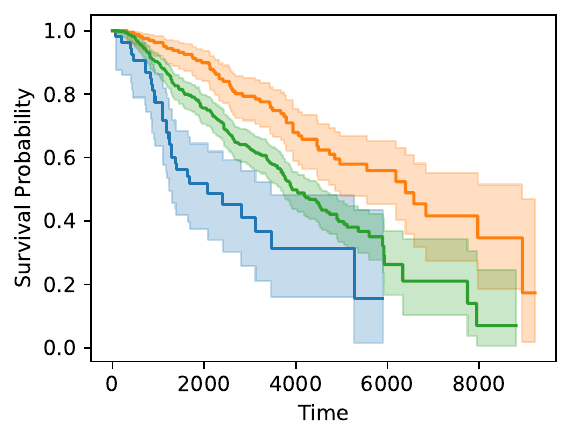}
\end{minipage}
\hfill
\begin{minipage}[c]{0.23\linewidth}
    \centering
    \includegraphics[width=\linewidth]{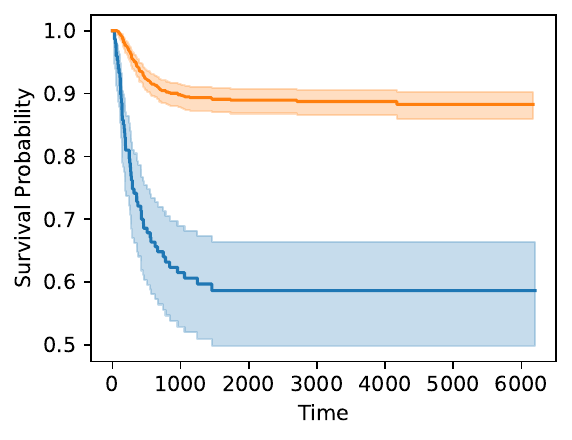}
\end{minipage}

\vspace{2mm}

%==================== ROW 2: K-Means ====================

\begin{minipage}[c]{0.04\linewidth}
    \centering
    \rotatebox{90}{\tiny\textbf{K-Means}}
\end{minipage}
\hfill
\begin{minipage}[c]{0.23\linewidth}
    \centering
    \includegraphics[width=\linewidth]{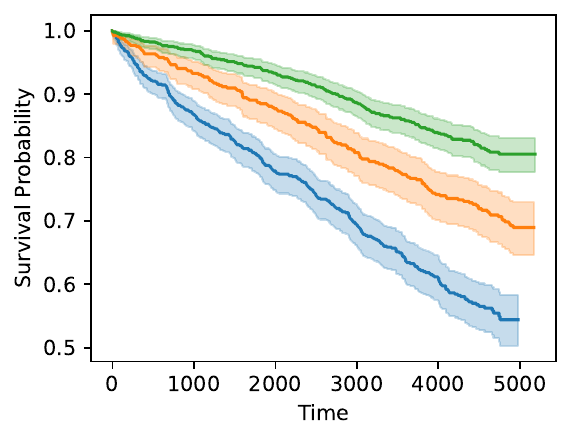}
\end{minipage}
\hfill
\begin{minipage}[c]{0.23\linewidth}
    \centering
    \includegraphics[width=\linewidth]{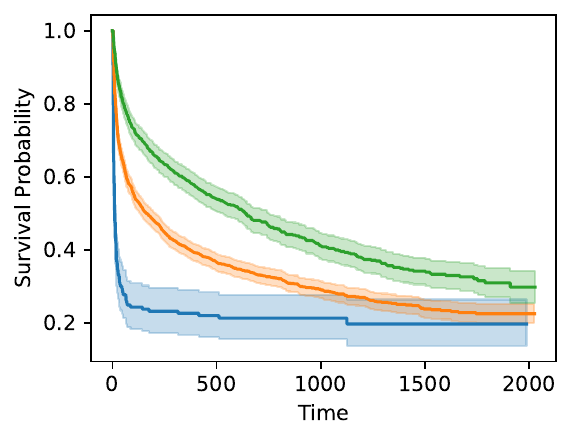}
\end{minipage}
\hfill
\begin{minipage}[c]{0.23\linewidth}
    \centering
    \includegraphics[width=\linewidth]{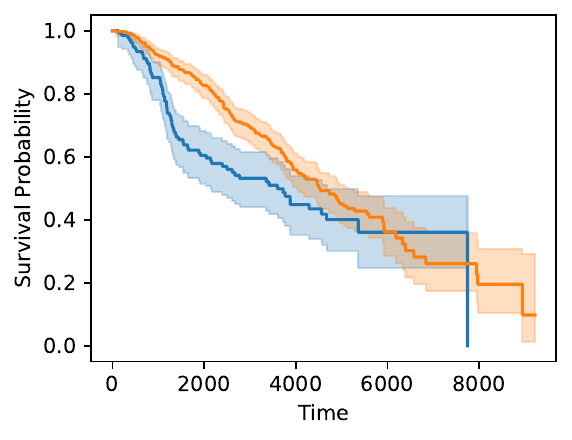}
\end{minipage}
\hfill
\begin{minipage}[c]{0.23\linewidth}
    \centering
    \includegraphics[width=\linewidth]{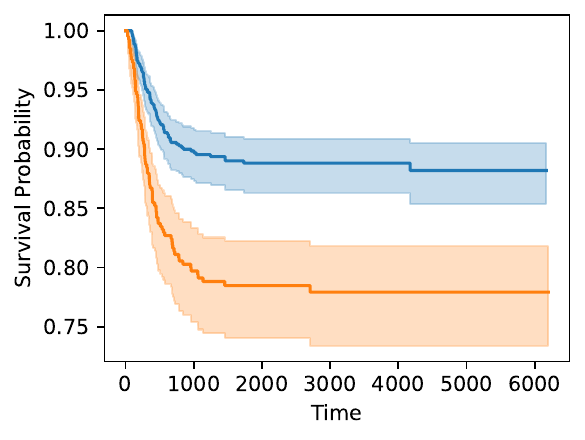}
\end{minipage}

\vspace{2mm}

%==================== ROW 2: Latent K-Means ====================

\begin{minipage}[c]{0.04\linewidth}
    \centering
    \rotatebox{90}{\tiny\textbf{K-Means (Latent)}}
\end{minipage}
\hfill
\begin{minipage}[c]{0.23\linewidth}
    \centering
    \includegraphics[width=\linewidth]{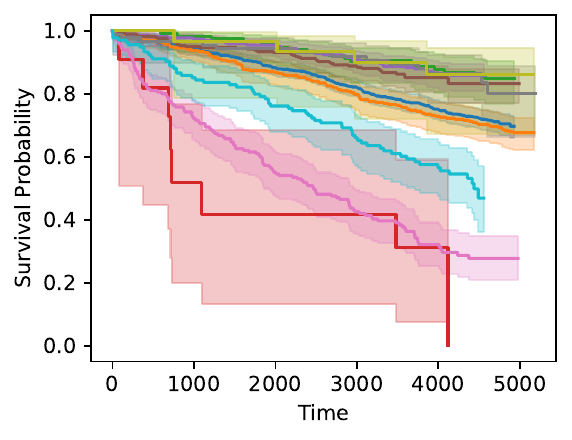}
\end{minipage}
\hfill
\begin{minipage}[c]{0.23\linewidth}
    \centering
    \includegraphics[width=\linewidth]{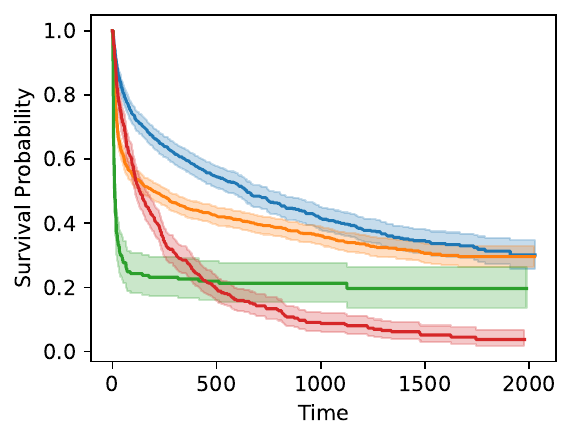}
\end{minipage}
\hfill
\begin{minipage}[c]{0.23\linewidth}
    \centering
    \includegraphics[width=\linewidth]{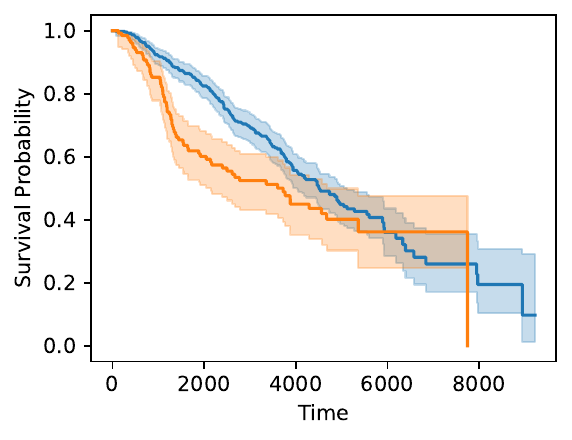}
\end{minipage}
\hfill
\begin{minipage}[c]{0.23\linewidth}
    \centering
    \includegraphics[width=\linewidth]{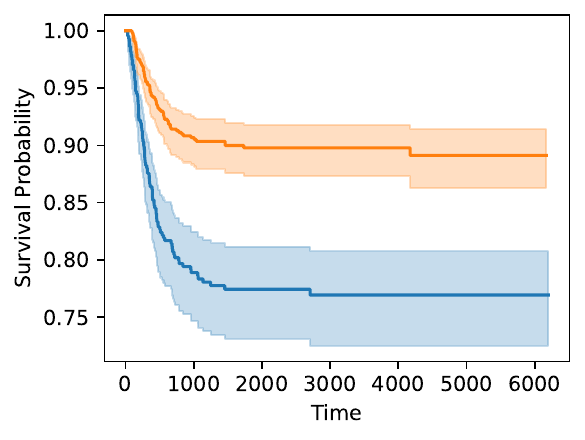}
\end{minipage}

\vspace{2mm}

%==================== ROW 3: SCA ====================

\begin{minipage}[c]{0.04\linewidth}
    \centering
    \rotatebox{90}{\tiny\textbf{SCA}}
\end{minipage}
\hfill
\begin{minipage}[c]{0.23\linewidth}
    \centering
    \includegraphics[width=\linewidth]{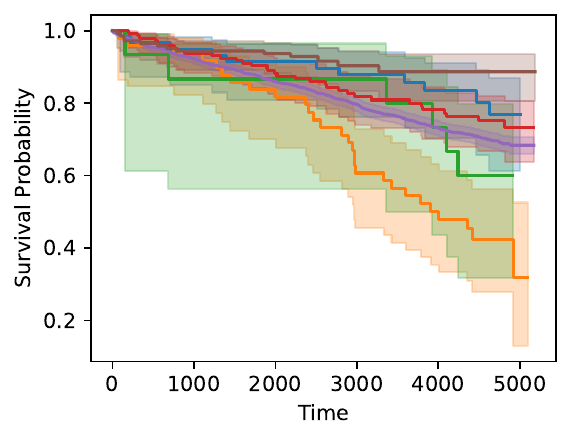}
\end{minipage}
\hfill
\begin{minipage}[c]{0.23\linewidth}
    \centering
    \includegraphics[width=\linewidth]{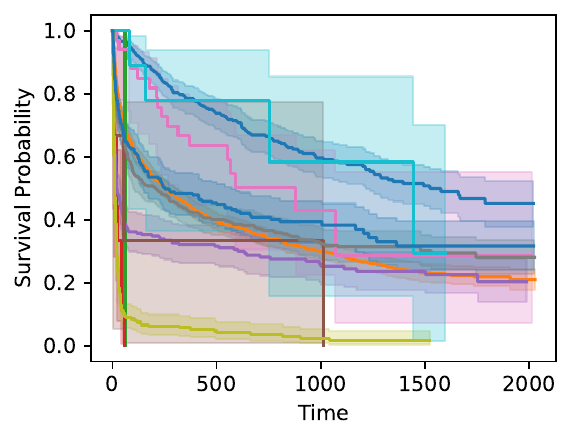}
\end{minipage}
\hfill
\begin{minipage}[c]{0.23\linewidth}
    \centering
    \includegraphics[width=\linewidth]{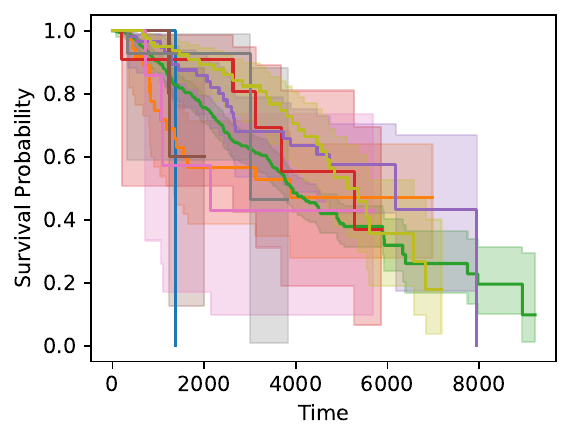}
\end{minipage}
\hfill
\begin{minipage}[c]{0.23\linewidth}
    \centering
    \includegraphics[width=\linewidth]{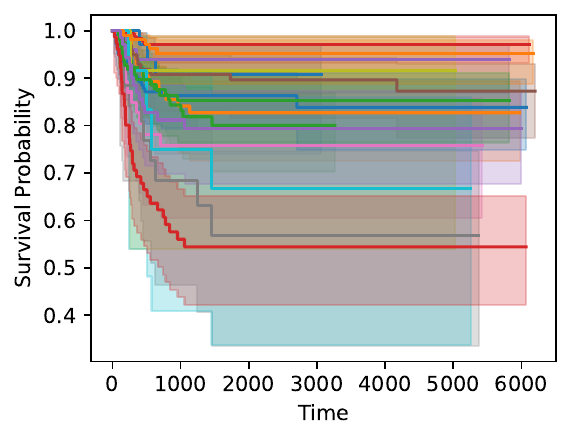}
\end{minipage}

\vspace{2mm}

%==================== ROW 4: VaDeSC ====================

\begin{minipage}[c]{0.04\linewidth}
    \centering
    \rotatebox{90}{\tiny\textbf{VaDeSC}}
\end{minipage}
\hfill
\begin{minipage}[c]{0.23\linewidth}
    \centering
    \includegraphics[width=\linewidth]{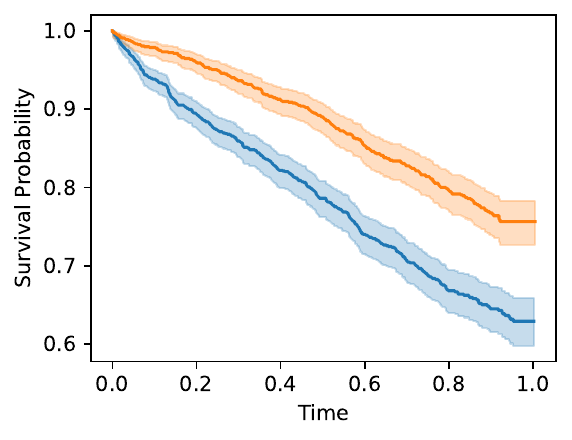}
\end{minipage}
\hfill
\begin{minipage}[c]{0.23\linewidth}
    \centering
    \includegraphics[width=\linewidth]{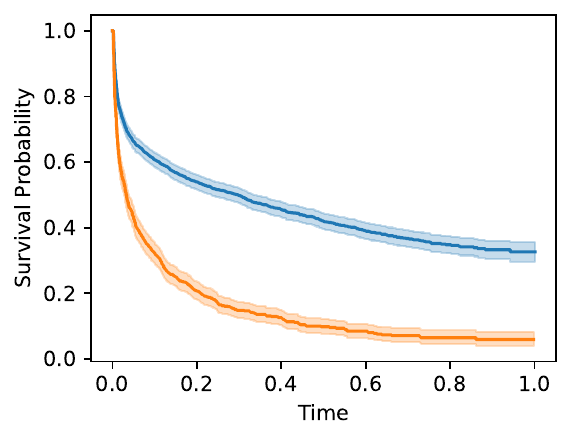}
\end{minipage}
\hfill
\begin{minipage}[c]{0.23\linewidth}
    \centering
    \includegraphics[width=\linewidth]{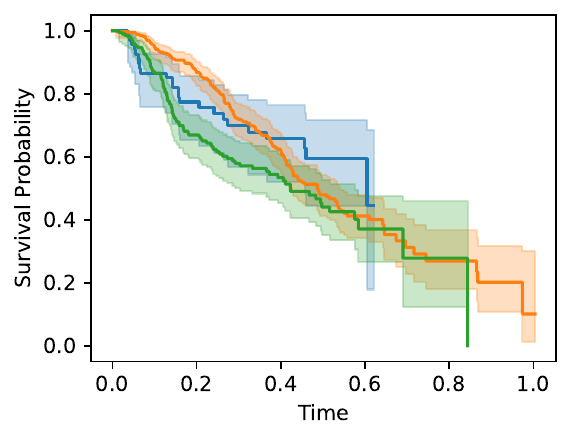}
\end{minipage}
\hfill
\begin{minipage}[c]{0.23\linewidth}
    \centering
    \includegraphics[width=\linewidth]{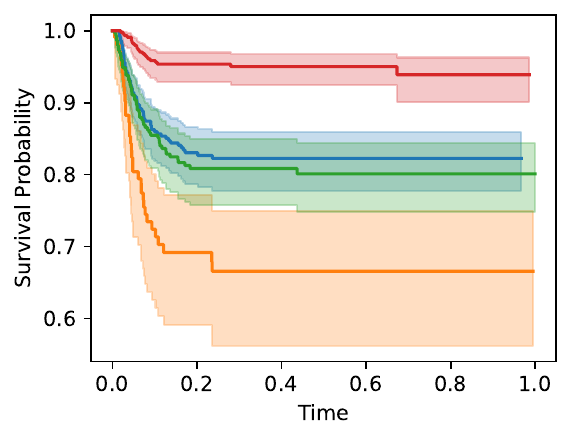}
\end{minipage}

\caption{Kaplan-Meier survival curves of the resulting clusters across datasets (columns) and clustering methods (rows).}
\label{fig:survival_curves}

\end{figure}

The good survival separation achieved by K-SurvMeans(Latent) is further supported by the Kaplan--Meier survival curves shown in Figure~\ref{fig:survival_curves}. The figure shows that across the datasets considered, K-SurvMeans(Latent) discovered more clusters than K-Means, and K-Means(Latent). Moreover, the discovered clusters exhibit clear differences with a limited overlap in survival curves compared to deep-learning-based models.

\section{Conclusion}
In this work, we introduced K-Survival Means (K-SurvMeans), an extension to the K-Means algorithm for survival analysis that incorporates the survival outcome into the optimization process to discover clusters with distinct survival patterns. The method benefits from using PSO to directly optimize the log-rank pairwise differences, thereby achieving a high percentage of significant differences between clusters. 

A Major limitation of the K-SurvMeans is its limited scalability to higher dimensions and to larger numbers of clusters. We showed that this limitation can be addressed by using a dimensionality reduction method, such as PCA, in this work. However, more advanced non-linear methods for dimensionality reduction can be utilized. Moreover, in this work, the dimensionality reduction is performed independently of the clustering method. While proven efficient, incorporating embedding learning into the optimization process can potentially improve results. 

Moreover, one key limitation of the proposed method compared to the two deep-learning approaches (SCA, and VaDeSC) is that, in addition to clustering, they predict individualized survival curves.

Extending the method to predict individualized survival curves and exploring different dimensionality-reduction and embedding-learning techniques are planned for future work.

\bibliographystyle{splncs04}
\bibliography{samplepaper}




% LNCS Homepage, \url{http://www.springer.com/lncs}, last accessed 2023/10/25
% \end{thebibliography}
\end{document}